\def\year{2018}\relax
\documentclass[letterpaper]{article} 
\usepackage{aaai18}  
\usepackage{times}  
\usepackage{helvet}  
\usepackage{courier}  
\usepackage{url}  
\usepackage{graphicx}  
\usepackage{algorithm}
\usepackage{algorithmic}
\usepackage{subfig}
\usepackage{amsmath}
\usepackage{amsfonts}
\usepackage{array}
\usepackage{todonotes}   
\usepackage{mathtools}
\begin{document}


%
\title{From Virtual Demonstration to Real-World Manipulation Using LSTM and MDN}
\author{Rouhollah Rahmatizadeh, Pooya Abolghasemi, Aman Behal, Ladislau B{\"o}l{\"o}ni\\
Department of Computer Science\\
University of Central Florida, 
United States\\
\texttt{rrahmati,pooya.abolghasemi,lboloni@cs.ucf.edu, abehal@ucf.edu} \\
}

\maketitle
\begin{abstract}
Robots assisting the disabled or elderly must perform complex manipulation tasks and must adapt to the home environment and preferences of their user. Learning from demonstration is a promising choice, that would allow the non-technical user to teach the robot different tasks. However, collecting demonstrations in the home environment of a disabled user is time consuming, disruptive to the comfort of the user, and presents safety challenges. It would be desirable to perform the demonstrations in a virtual environment. 

In this paper we describe a solution to the challenging problem of behavior transfer from virtual demonstration to a physical robot. The virtual demonstrations are used to train a deep neural network based controller, which is using a Long Short Term Memory (LSTM) recurrent neural network to generate trajectories. The training process uses a Mixture Density Network (MDN) to calculate an error signal suitable for the multimodal nature of demonstrations. The controller learned in the virtual environment is transferred to a physical robot (a Rethink Robotics Baxter). An off-the-shelf vision component is used to substitute for geometric knowledge available in the simulation and an inverse kinematics module is used to allow the Baxter to enact the trajectory. 

Our experimental studies validate the three contributions of the paper: (1) the controller learned from virtual demonstrations can be used to successfully perform the manipulation tasks on a physical robot, (2) the LSTM+MDN architectural choice outperforms other choices, such as the use of feedforward networks and mean-squared error based training signals and (3) allowing imperfect demonstrations in the training set also allows the controller to learn how to correct its manipulation mistakes.
\end{abstract}



\section{Introduction}
\label{sec:Introduction}

Assistive robotics, whether in the form of wheelchair mounted robotic arms or mobile robots with manipulators, promises to improve the independence and quality of life of the disabled and the elderly. Such robots can help users to perform Activities of Daily Living (ADLs) such as self-feeding, dressing, grooming, personal hygiene and leisure activities. Almost all ADLs involve some type of object manipulation. While most current systems rely on remote control, there are ongoing research efforts to make assistive robots more autonomous~\cite{endres2013learning,miller2012geometric,bollini2011bakebot}. One of the challenges of assistive robotics is the uniqueness of every home and the need to adapt to the preferences and disabilities of the user. As the user is a non-programmer, learning from demonstration (LfD) is a promising way to adapt the robot behavior to the specific environment, objects and preferences. However, collecting large numbers of physical demonstrations from a disabled user is challenging. For instance, the manipulated objects and the environment can be fragile and the wheelchair-bound users might have difficulty recovering from a failed manipulation task such as a dropped cup. Similar considerations hinder from-the-scratch reinforcement learning performed in a home environment.

\begin{figure*}
  \centering
  \includegraphics[width=1\textwidth]{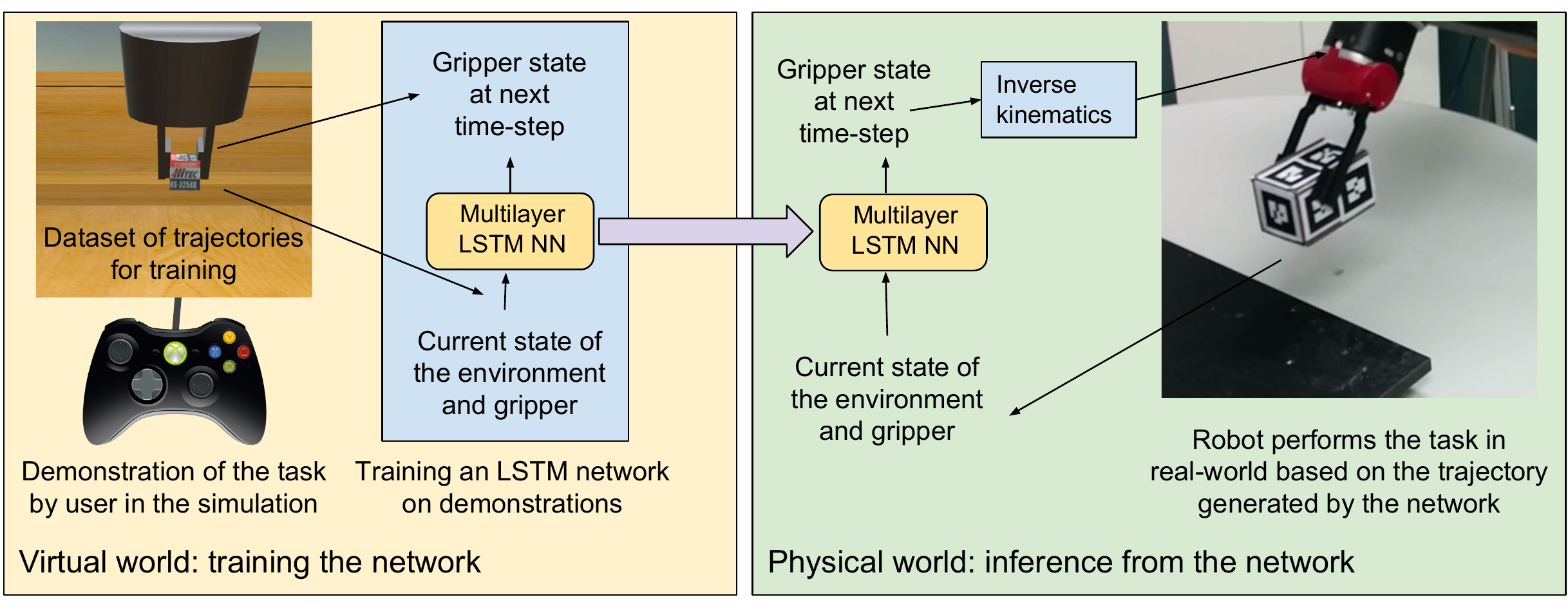}
  \caption{The general flow of our approach. The demonstrations of the ADL manipulation tasks are collected in a virtual environment. The collected trajectories are used to train the neural network controller. }
  \label{fig:flow}
\end{figure*}

In this paper we propose an approach where the users demonstrate the tasks to be performed in a virtual environment. This allows the safe collection of sufficient demonstrations to train a deep neural network based robot controller. The trained controller is then transferred to the physical robot. The general flow is illustrated in Figure~\ref{fig:flow}.  

In the remainder of this paper we discuss related work, describe the approach in detail, and through an experimental study validate the three contributions outlined in the abstract.

\section{Related work}

\label{sec:RelatedWork}

\noindent{\em Virtual training to physical execution.} The desirability of transferring learning from a simulated robot to a physical one had been recognized by many researchers. This need is especially strong in the case of reinforcement learning-based approaches, where the exploration phase can be very expensive on a physical robot. Unfortunately, policies learned in simulation do not work well when transferred na{\"i}vely to a physical robot. First, no simulation can be perfect, thus the physical robot will inevitably have a suboptimal policy leaving to a slightly different state from the expected one. This would be acceptable if the policy corrects the error, keeping the difference bounded. In practice, it was found that if a policy was learned from demonstrations, initially small differences between the simulation and the real world tend to grow larger and larger as the state diverges farther and farther from the settings in which the demonstrations took place -- an aggravated version of the problem that led to the development of algorithms such as DAgger~\cite{ross2011reduction}.


A number of different approaches had been developed to deal with this problem. One approach is to try to bring the simulation closer to reality through learning. This approach had been taken by Grounded Simulation Learning~\cite{farchy2013humanoid} and improved by Grounded Action Transformation~\cite{hanna2017grounded} on the task of teaching a Nao robot to walk faster. 

Another approach for improving the virtual to physical transfer is to increase the generality of the policy learned in simulation through domain randomization~\cite{tobin2017domain}. \cite{christiano2016transfer} note that the sequence of states learned by the controller in simulation might be reasonable even if the exact controls in the physical world are different. Their approach computes what the next state would be, and relies on learned deep inverse dynamics model to decide on the actions to achieve the equivalent real world state. \cite{james2017transferring} train a CNN+LSTM controller to perform a multi-step task of picking up a cube and dropping it to a table. The training is based on demonstrations acquired from a programmed optimal controller, with domain randomization in the form of  variation of environmental characteristics, lighting and textures. 

Yet another possible approach is to learn an invariant feature space that can transfer information between the domains through a sort of analogy making~\cite{gupta2017learning}.

In the work described in this paper, we took a different approach. We had to accept that the simulation is nowhere close to the physical environment, thus the robot {\em will} make mistakes. Instead of domain randomization, we rely on the natural imperfections of demonstrations done by (possibly disabled) humans, but also on the ability of humans to correct the errors they made. Using this training data, we found that the physical robot also learned to correct its mistakes.

\medskip

\noindent{\em Autonomous trajectory execution through Learning from Demonstration (LfD).} LfD  extracts policies from example state to action mappings~\cite{argall2009survey} demonstrated by a user. Successful LfD applications include autonomous helicopter maneuvers~\cite{abbeel2010autonomous}, playing table tennis~\cite{kober2011reinforcement,calinon2010learning}, object manipulation~\cite{pastor2009learning}, and making coffee~\cite{sung_robobarista_2015}.

The greatest challenge of LfD is generalization to unseen situations. One obvious way to mitigate this problem is by acquiring a large number of demonstrations covering as many situations as possible. Some researchers proposed cloud-based and crowdsourced data collection techniques~\cite{kehoe2013cloud,forbes2014robot,crick2011human}, and some others proposed to use simulation environments~\cite{fang2016learning}. Another direction is to use smaller number of demonstrations, but change the learning model to generalize better. One possible technique is to hand-engineer task-specific features~\cite{calinon2007learning,calinon2009handling}. Using deep neural networks might help to eliminate the need to hand-engineer features. For instance,~\cite{levine2016end} used feed-forward neural networks to map a robot's visual input to control commands. The visual input is processed using a convolutional neural network (CNN) to extract 2D feature points, then it is aggregated with the robot's current joint configuration, and fed into a feed-forward neural network. The resulting neural network will predict the next joint configuration of the robot. A similar neural network architecture is designed to address the grasping problem~\cite{pinto2015supersizing}. In~\cite{mayer2008system}, LSTMs are used for a robot to learn to autonomously tie knots after a pre-processing step to reduce the noise from the training data.

\section{Our approach}

\subsection{Collecting demonstrations in a virtual environment}

To allow users to demonstrate their ADLs, we designed in the Unity3D game engine a virtual environment modeling a table with an attached shelf that can hold various objects, and a simple two-finger gripper that can be opened and closed to grasp and carry an object. The user can use the mouse and keyboard to open/close the gripper, as well as to move and rotate it, giving it 7 degrees of freedom. Unity3D simulates the basic physics of the real world including gravity, collisions between objects as well as friction, but there is no guarantee that these will accurately match the real world. 

We represent the state of the virtual environment as the collection of the poses of the $M$ movable objects $q = \{o_1 \ldots o_M\}$. The pose of an object is represented by the vector containing the position and rotation quaternion with respect to the origin $o = [p_x, p_y, p_z, r_x, r_y, r_z, r_w]$. During each step of a demonstration, at time step $t$ we record the state of the environment $q_t$ and the pose of the end-effector augmented with the open/close status of the gripper $e_t$. Thus a full demonstration can be recorded as a list of pairs $d = \{ (q_1,e_1) \ldots (q_T,e_T)\}$.

For our experiments we considered two manipulation tasks that are regularly found as components of ADLs: pick and place and pushing to a desired pose.

The {\em pick and place} task involves picking up a small box located on top of the table, and placing it into a shelf above the table. The robot needs to move the gripper from its initial random position to a point close to the box, open the gripper, position the fingers around the box, close the gripper, move towards the shelf in an orientation where it will not collide with the shelf, enter the shelf, and finally open the gripper to release the box. This task is a clearly segmented multi-step task, where the steps need to be executed in a particular order. On the other hand, the execution of the task depends very little on the details of the physics: as long as the box can be picked up by the gripper, its weight or friction does not matter. 

The {\em pushing to desired pose} task involves moving and rotating a box of size $10 \times 7 \times 7$cm to a desired area only by pushing it on the tabletop. In this task, the robot is not allowed to grasp the object. The box is initially positioned in a way that needs to be rotated by $90^{\circ}$ to fit inside the desired area which is $3$cm wider than the box in each direction. The robot starts from an initial gripper position, moves the gripper close to the box and pushes the box at specific points at its sides to rotate and move it. If necessary, the gripper needs to circle around the box to find the next contact point. This task does not have a clearly sequenced set of steps. It is unclear how many times does the box needs to be pushed. Furthermore, the completion of the task depends on the physics: the weight of the box and the friction between the box and table impacts the way the box moves in response to pushes.

  
The demonstrations were collected from a single user, in the course of multiple sessions. In each session, the user performed a series of demonstrations for each task. The quality of demonstrations varied: in some of them, the user could finish the task only after several tries. For instance, sometimes the grasp was unsuccessful, or the user dropped the object in an incorrect position and had to pick it up again. After finishing a demonstration, the user was immediately presented with a new instance of the problem, with randomly generated initial conditions. All the experiments had been recorded in the trajectory representation format presented above, at a recording frequency of 33Hz. However, we found that the neural network controller can be trained more efficiently if the trajectories are sampled at a lower rate, with a rate of 4Hz to giving the best results.

To improve learning, we extended our training data by exploiting both the properties of the individual tasks and trajectory recording technique. First, we noticed that in the pick and place task the user can put the object to any location on the shelf. Thus we were able to generate new synthetic training data by shifting the existing demonstration trajectories parallel with the shelf. As the pushing to desired pose task requires a specific coordinate and pose to succeed, this approach is not possible for the second task.

\begin{figure}[h]
  \centering
  \includegraphics[width=\columnwidth]{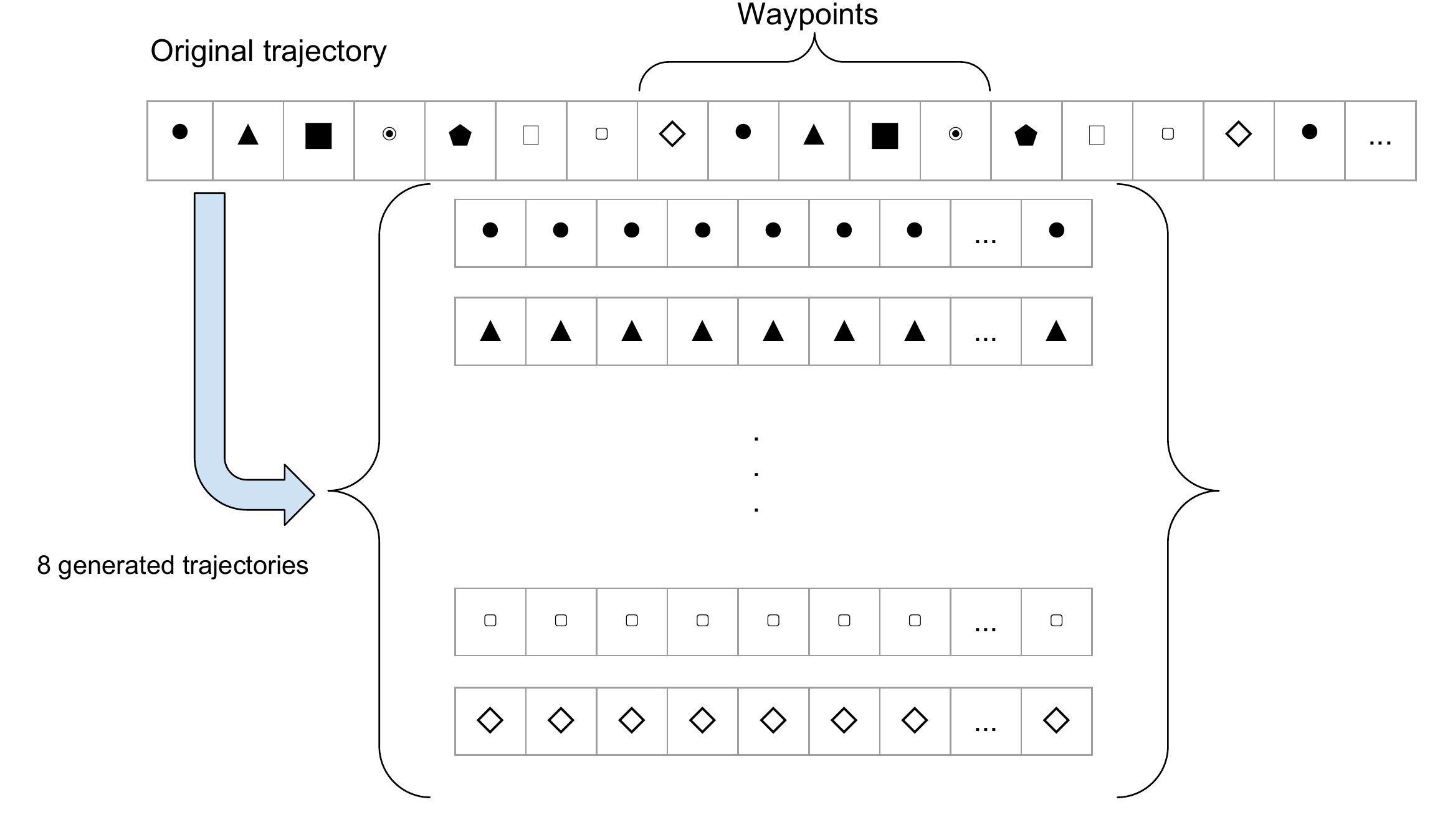}
  \caption{Creating multiple trajectories from a demonstration recorded at a higher frequency.}
  \label{fig:frequencyReduction}
\end{figure}

The second observation was that by recording the demonstration at 33Hz but presenting the training trajectories at only 4Hz, we have extra trajectory points. These trajectory points can be used to generate multiple independent trajectories at a lower temporal resolution. The process of the trajectory generation by frequency reduction is shown in Figure~\ref{fig:frequencyReduction}. Table~\ref{Dataset} describes the size of the final dataset. 

\begin{table}[h]
\centering
{\footnotesize
\begin{tabular}{p{4.5cm}p{1.2cm}p{1cm}}
\hline \\[-.8em]
Task & Pick and place & Push to pose \\
\\[-.8em]
\hline
\\[-.8em]
 Raw demonstrations & 650 & 1614  \\
Demonstrations after shift & 3900 & - \\
Low frequency demonstrations & 31,200 & 12,912 \\
 Total no. of waypoints  & 645,198 & 369,477 \\
 Avg. demonstration waypoints & 20.68 & 28.61 \\ \\[-.8em]
\hline 
\\[-.8em]
\end{tabular}
}
\caption{The size of the datasets for the two studied tasks \label{Dataset}}
\end{table}

\subsection{The neural network based robot controller}

The robot controller takes as input the pose of the objects involved and the pose and open/close status of the gripper at time $t$ and outputs a prediction of the pose and the open/closed status of the gripper at time $t+1$. During training, this prediction is used to generate the error signal. During the deployment of the trained network, the prediction represents the desired pose of the end actuator which the robot needs to achieve through its inverse kinematics calculations. The controllers for the ``pick and place'' and the ``push to desired pose'' tasks have the same network architecture but were trained on the specific tasks. 

Our architecture uses an LSTM recurrent neural network and relies on mixture density networks (MDNs) to predict the probability density of the output. The error signal, in this case, is based on the negative logarithm likelihood of the next target waypoint given the probability density implied by the MDN.

\begin{figure*}
  \centering
  \includegraphics[width=1\textwidth]{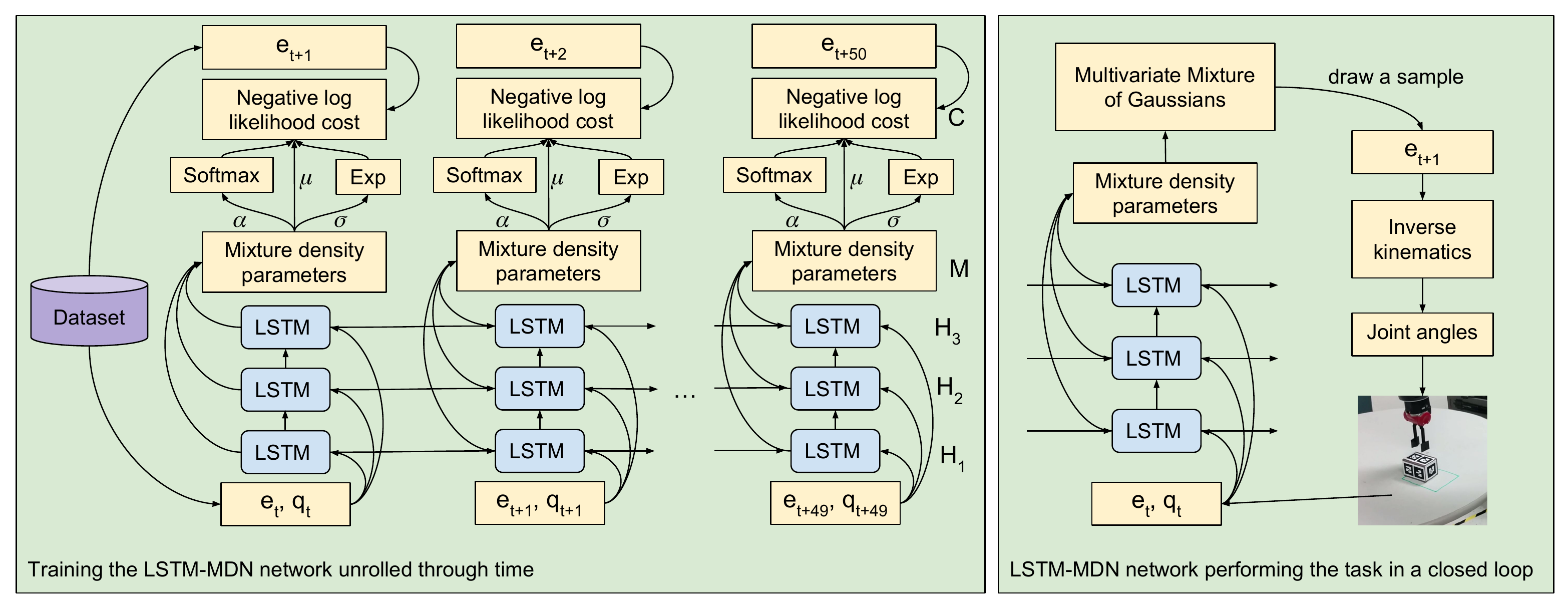}
  \caption{The training and evaluation phase. During the training the LSTM network is unrolled for 50 time-steps. The gripper pose and status (open/close) $e_t$ and the pose of relevant objects $q_t$ at time-step $t$ is used as input and output of the network to calculate and backpropagate the error to update the weights. During the evaluation phase, the mixture density parameters are used to form a mixture of Gaussians and draw a sample from it. The sample is used to control the robot arm. }
  \label{fig:network_structure}
\end{figure*}

Let us now describe the intuitions that led to these choices. The solution to both manipulation tasks contain a series of individual movements which need to be executed in a specific sequence. Although both tasks can be solved in several different ways, the individual movements in them cannot be randomly exchanged. In order to successfully solve the task, the robot needs to choose and {\em commit to a certain solution}. While it is technically possible that this commitment will be encoded in the environment outside the robot, we conjecture that a robot controller which has a memory to store these commitments will perform better. The requirement of a controller with a memory leads us to the choice of recurrent neural networks, in particular, one of the most widely used models, the LSTM~\cite{hochreiter1997long}. We are using three LSTM layers with 50 nodes each as shown in Figure~\ref{fig:network_structure}.

The second intuition applies to the choice of the output layer and error signal. Both tasks allow multiple solutions. For instance, for the push to pose task, the robot might need to move the box in a diagonal direction. This can be achieved by either (a) first pushing the shorter side of the box followed by a push on the longer side or (b) the other way around. However, by averaging these two solutions we reach a solution where the gripper would try to push the corner of the box, leading to an unpredictable result. This leads us to conjecture that a multi-modal error function would perform better than the unimodal MSE. The approach we chose is based on Mixture Density Networks (MDN)~\cite{bishop1994mixture} which is to use the output of the network to predict the parameters of a mixture distribution. The output of the network (which, in our case, is the next pose of the gripper), will be a sample drawn from this distribution. Unlike the model with MSE cost which is deterministic, this approach can model stochastic behaviors to be executed by the robot. The probability density of the next waypoint can be modeled using a linear combination of Gaussian kernel functions
\begin{equation}
p(y|x) = \sum_{i=1}^{m} \alpha_i(x) g_i(y|x)
\end{equation}
where $\alpha_i(x)$ is the mixing coefficient, $g_i(y|x)$ is a multivariate Gaussian, and $m$ is the number of kernels. Note that both the mixing coefficients and the Gaussian kernels are conditioned on the complete history of the inputs till current timestep $x = \{x_1 \ldots x_t\}$. This is because the concatenation of the output of all layers which is used to estimate the mixing coefficients and Gaussian kernels is itself a function of $x$. The Gaussian kernel is of the form
\begin{equation}
g(y|x) = \frac{1}{{ {(2\pi)}^{c/2}\sigma_i(x) }}\exp\left \{{{ - \frac{\|y - \mu_i(x)\|^2}{2\sigma_i(x)^2}}}\right \}
\end{equation}
where the vector $\mu_i(x)$ is the center of $i$th kernel. We do not calculate the full covariance matrices for each component, since this form of Gaussian mixture model is general enough to approximate any density function~\cite{mclachlan1988mixture}.

The parameters of the Gaussian kernels $\mu_i(x)$, $\sigma_i(x)$ and mixing coefficients $\alpha_i(x)$ are represented by the layer M in Figure~\ref{fig:network_structure}. To accomplish this, layer M needs to have one neuron for each parameter. Thus layer M will have a width of $(c+2)\times m$, containing $c\times m$ neurons for $\mu_i(x)$, $m$ neurons for $\sigma_i(x)$, and another $m$ neurons for $\alpha_i(x)$. This layer is fully connected to the concatenation of layers $H_1$, $H_2$ and $H_3$.

To satisfy the constraint $\sum_{i=1}^{m} \alpha_i(x) = 1$, the corresponding neurons are passed through a softmax function. The neurons corresponding to the variances $\sigma_i(x)$ are passed through an exponential function and the neurons corresponding to the means $\mu_i(x)$ are used without any further changes. Finally, we can define the error in terms of negative logarithm likelihood
\begin{equation}
E_\mathit{MDN} = -ln \left \{ \sum_{i=1}^{m} \alpha_i(x) g_i(y|x) \right \}
\end{equation}

The network is unrolled for 50 time steps. All the parameters are initialized uniformly between -0.08 to 0.08 following the recommendation by \cite{sutskever2014sequence}. Stochastic gradient descent with mini-batches of size 10 is used to train the network. RMSProp with initial learning rate of 0.001 and decay of 0.99-0.999 (based on number of examples) is used to divide the gradients by a running average of their recent magnitude. In order to overcome the exploding gradients problem, the gradients are clipped in the range [-1, 1]. We use 80\% of the data for training and keep the remaining 20\% for validation. We stop the training when the validation error does not change for 20 epochs.


\subsection{Transferring the controller to the physical robot}

The last step of the process is to transfer the trained controller to a physical robot (a Rethink Robotics Baxter). As the controller provides only the next pose of the end effector, the controller had been augmented with an inverse kinematics module to calculate the trajectory in the Baxter robot arms' joint space. 

Another challenge is that while in the virtual world we had perfect knowledge of the pose of the effector and all the objects in the environment, we needed to acquire this information through sensing. As our controller architecture only performs robot arm control, in order to supplant the missing vision component, we relied on a Microsoft Kinect sensor and objects annotated with markers to track their pose. One of the problems with this approach is that the robot arm might occlude the view of the sensor. The Kinect sensor was placed close to the table to reduce the chance of occlusion, however, occlusions may still occur if the robot's arm is placed between the object and the Kinect. Another challenge is the fact that the waypoints generated by the controller are relatively far away, leading to a jerky motion. To smooth the robot's movement we use interpolation in joint space to fill in the gap between the current waypoint and the previous one. 

Finally, we found that the trajectory described by the controller can not always be executed by the Baxter arm at the same speed as in the virtual environment. Therefore, we use a dynamic execution rate to wait between execution of each waypoint. Concretely, the algorithm waits for 0.2sec and checks if the difference between the current pose of the gripper and the predicted one is below a certain threshold. If yes, it commands the robot to go to the next waypoint, otherwise it waits in a loop until the end-effector reaches the desired pose or timeout occurs which means that the end-effector cannot reach that pose (either because the inverse kinematic failed or a collision occurred).

\section{Experimental study}
\label{sec:Experiments}

In the following we describe a series of experiments and observations justifying the three claims we made in the abstract of this paper.

\begin{figure*}
	\centering
    \includegraphics[height=.86in]{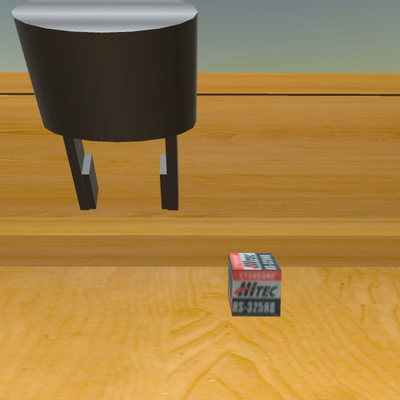}
    \includegraphics[height=.86in]{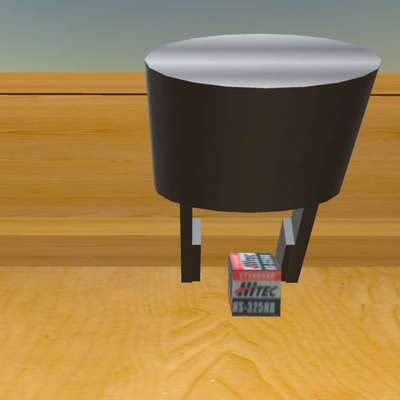}
    \includegraphics[height=.86in]{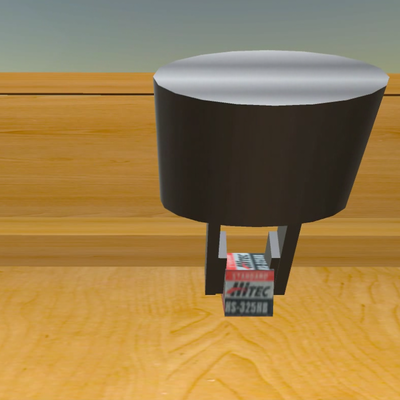}
    \includegraphics[height=.86in]{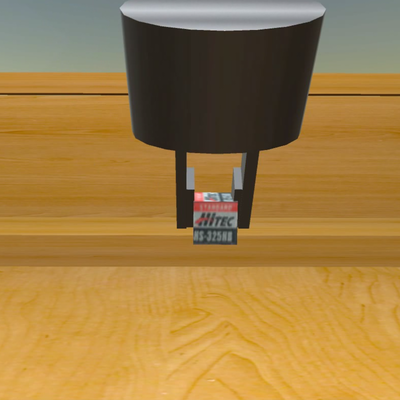}
    \includegraphics[height=.86in]{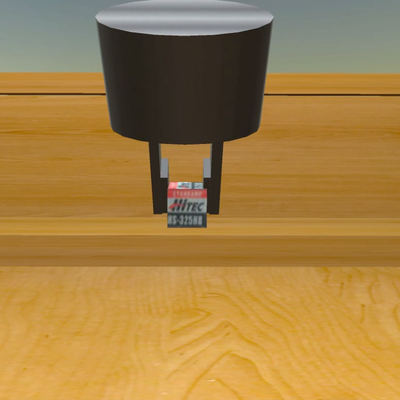}
    \includegraphics[height=.86in]{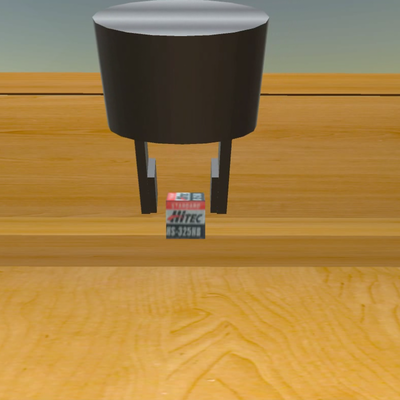}
    \newline
    \includegraphics[height=.86in]{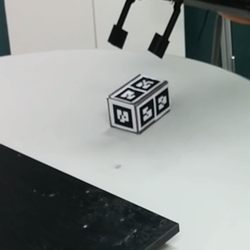}
    \includegraphics[height=.86in]{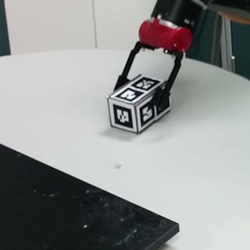}
    \includegraphics[height=.86in]{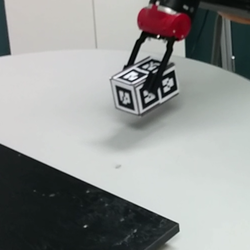}
    \includegraphics[height=.86in]{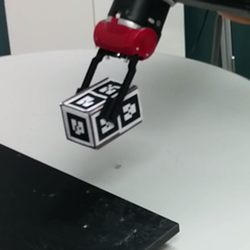}
    \includegraphics[height=.86in]{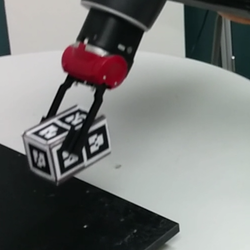}
    \includegraphics[height=.86in]{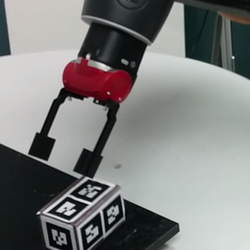}
    \newline
    \includegraphics[height=.86in]{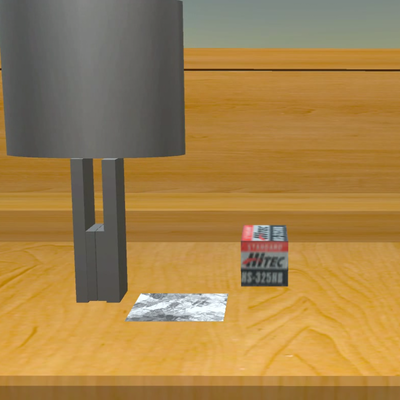}
    \includegraphics[height=.86in]{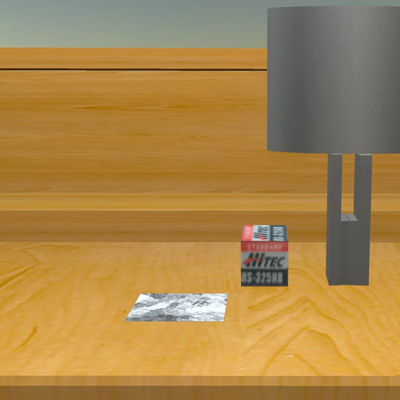}
    \includegraphics[height=.86in]{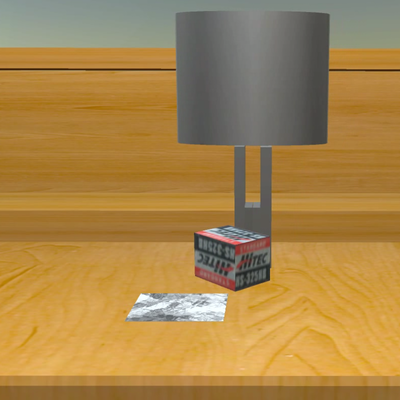}
    \includegraphics[height=.86in]{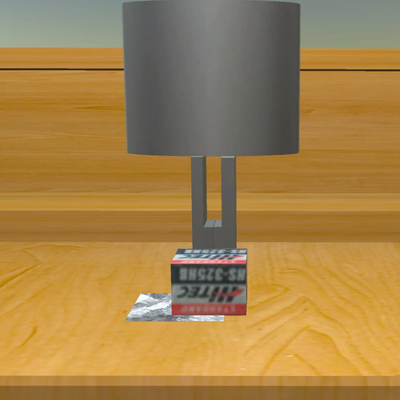}
    \includegraphics[height=.86in]{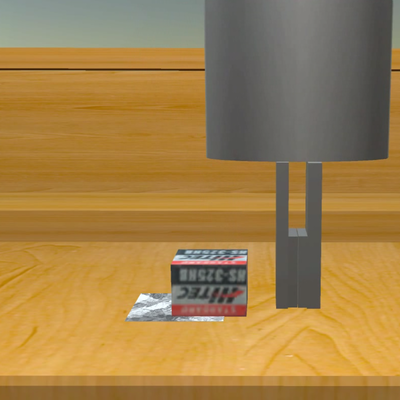}
    \includegraphics[height=.86in]{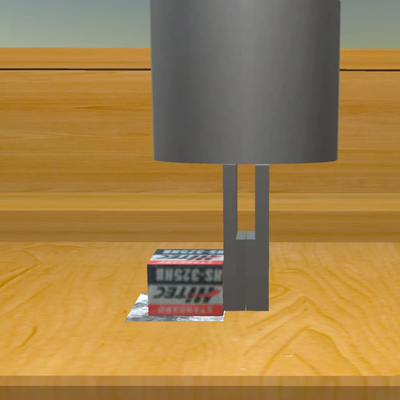}
    \newline
    \includegraphics[height=.86in]{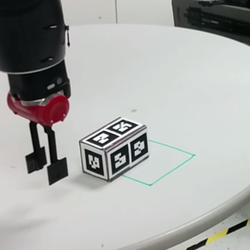}
    \includegraphics[height=.86in]{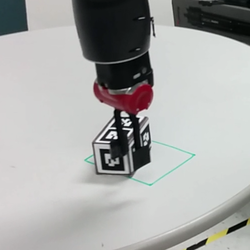}
    \includegraphics[height=.86in]{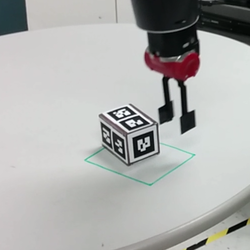}
    \includegraphics[height=.86in]{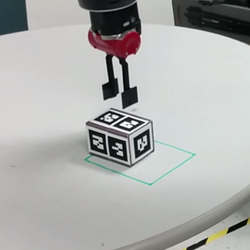}
    \includegraphics[height=.86in]{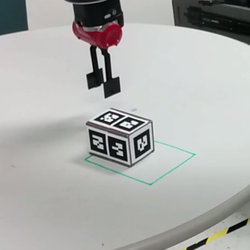}
    \includegraphics[height=.86in]{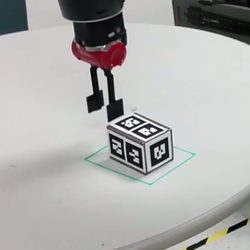}
    \newline
	\caption{A sequence of images showing the autonomous execution of pick and place in simulation (first row), pick and place in real world (second row), pushing in simulation (third row), and pushing in real world (fourth row). The robot is controlled by a mixture density network with 3 layers of LSTM.}
	\label{fig:task_execution}

\end{figure*}

\subsection{The benefits of LSTM and MDN}

Our network architecture uses LSTM layers and an MDN-based error signal. These choices need to be validated, as many researchers reported success on similar tasks with a simpler network - without recurrent layers and using mean squared error (MSE) as an error signal. For instance,~\cite{levine2016end} use convolutional layers (for extracting poses from images) followed by feedforward layers to give a unimodal prediction about the next waypoint in the trajectory. As our controller is not end-to-end (in real world experiments we are using an off-the-shelf solution for computer vision), we can only compare our controller with the part of the controller from~\cite{levine2016end} that follows the convolutional layers used for image processing. In order to be able to perform a rough comparison, we implemented a feedforward network that closely matches that controller, but replaces the convolutional layers with a direct position input. 

With this, we have four choices for the controller structure: 

{\em FeedForward-MSE:} 3 layers of fully connected feedforward network with 100 neurons in each layer and mean squared error as the cost function.

{\em LSTM-MSE:} 3 layers of LSTM with 50 memory states in each layer and mean squared error as the cost function.
  
{\em FeedForward-MDN:} Mixture density network containing 3 fully connected feedforward layers with 100 neurons in each layer. The mixture contains 20 Gaussian kernels.

{\em LSTM-MDN:} Mixture density network containing 3 layers of LSTM with 50 memory states in each layer. The mixture contains 20 Gaussian kernels.

\begin{figure}[!t]
  \centering
  \includegraphics[width=0.80\columnwidth]{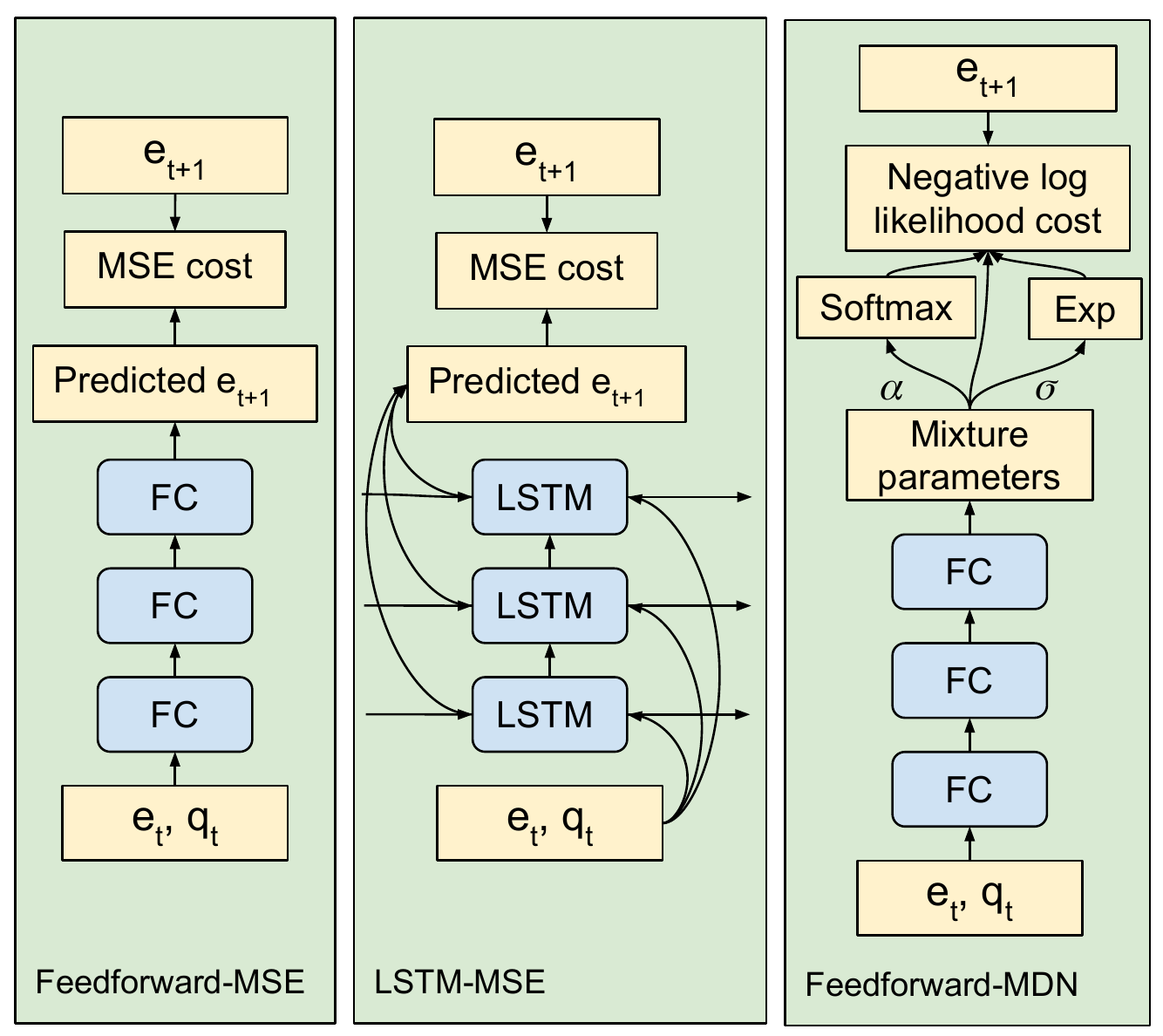}
  \caption{Alternative network architectures used in the comparison study: Feedforward-MSE, LSTM-MSE and Feedforward-MDN}
  \label{fig:network_comparison}
\end{figure}

The LSTM-MDN network is described in Figure~\ref{fig:network_structure}, while the architectures of the other approaches are shown in Figure~\ref{fig:network_comparison}. Each network had been separately trained for the pick and place and the push to desired pose respectively, in effect creating 8 different controllers. The resulting controllers had been tested in the virtual environment by requiring the robot to perform randomly generated tasks 20 times. If it can not complete the task in a limited time (1 minute for the first task and 2 minutes for the second one), we count the try as a failure and reset the position of the box. The numerical success rates are shown in the following table:

 {\footnotesize
\begin{center}
\begin{tabular}{
p{3.0cm}cc } 
 \hline \\[-.8em]
 \textbf{Controller} & \textbf{Pick and place} & \textbf{Push to pose} \\ \\[-.8em]
 \hline \\[-.8em]
Feedfoward-MSE & 0\% & 0\% \\ 
LSTM-MSE & 85\% & 0\% \\ 
Feedforward-MDN  & 95\% & 15\% \\ 
 LSTM-MDN & \textbf{100\%} & \textbf{95\%}\\ 
 \hline
\end{tabular} 
\end{center}
}

The results allow us to derive several conclusions. Clearly, the Feedforward-MSE combination does not work for this particular set of problems and training data, failing to complete either task even once. Another conclusion is that the pick-and-place task is clearly the easier from the two, even when tested in the virtual environment where physical modeling errors don't play a role. The presence of either LSTM or MSE in the network provided enough improvement in the controller to allow the finishing of the task in the majority of situations, while the presence of both lead to 100\% success rate. On the other hand, the harder push to pose task requires both components to have a reasonable success rate of 95\%.

We conclude from these experiments that both adding LSTM layers, and using an MDN output provide significant benefits individually to the robot controller, and that these two techniques can be combined for more individual benefits.

\subsection{Evaluating the transfer to the physical robot}

To verify the ability of the controller trained in the virtual world to perform in the physical one, we subjected both the virtual and the physical robots to the same tasks. The sequence of images in the Figure~\ref{fig:task_execution} shows the controller acting autonomously for the pick and place and pushing to pose tasks in the virtual and physical environments respectively. We have found that indeed, in most cases, the physical robot was successful in executing both tasks. This is not because the virtual and physical worlds are highly similar. The size of the gripper of the Baxter robot is different from the one in the virtual world. The friction coefficients are very different, and the physics simulation in the virtual world is also of limited accuracy. Even the size and shape of the box used in the physical experiments is not an exact match of the ones in the simulation, and the physical setup suffered from camera calibration problems. Overall, the number of things that can go wrong, is much higher in the physical world. 

What we found remarkable is that, in fact, during the experiments many things went wrong or changed from the training data - however, the robot was often able to recover from them. This shows how the model is robust in handling deviation from what it has seen. The network contains different solutions for a case that can apply if others fail. For instance, in the pushing task, to rotate the box, it tries to touch the corner of the box and push it. If the box did not move since the gripper passed it without a touch, it tries again but this time from a point closer to the center of the box. This gives the network some tolerance to slight variations in the size of the box from the simulation to the real world.

To compare the success ratio in the virtual and physical world, we repeated the 20 experiments on the physical world, using the LSTM-MDN controller (the only one that had significant success rate on the virtual robot on both tasks). The success rates in the virtual and physical worlds are compared as follows:

{\footnotesize
\begin{center}
\begin{tabular}{
p{2.5cm}cc } 
 \hline \\[-.8em]
 \textbf{Environment} &  \textbf{Pick and place} & \textbf{Push to pose} \\ \\[-.8em]
 \hline
 \\[-.8em]
 Virtual world & 100\% & 95\%\\ 

 Physical world & 80\% & 60\%\\ 
  \hline
\end{tabular}
\end{center}
 }

The first conclusion we can draw from these values is that the approach successfully demonstrated the ability to transfer an unchanged controller trained in the virtual world to a physical robot. As expected, the success rate was lower in the physical world for both tasks. Some of the reasons behind the lower success rate is obvious: for instance, in the physical world there is an inevitable noise in the position of the objects and the end effector. Some of the noise is a consequence of limited sensor accuracy (such as the calibration of the Kinect sensor) and effector performance. Another source of inaccuracy is due to the way in which we acquired the positional information through a Kinect sensor: if during the manipulation the robot arm occluded the view of the object to the Kinect sensor, we temporarily lost the ability to track the object. 

Another reason for the lower performance in the physical world is due to the differences in the size, shape, physical attributes such as friction, etc. of the gripper and objects between the simulation and real world. As we discussed when introducing the problems, the push-to-pose tasks is more dependent of the physics (such as the friction between the object and the table determines the way the object moves when pushed). This creates a bigger difference between the virtual and the physical environment compared to the pick and place task, where after a successful grasp the robot is essentially in control of the environment. Thus, the push to pose task shows a stronger decrease in success rate when moving to the physical world. 

\subsection{Imperfect demonstrations teach the robot how to self-correct}

The conventional wisdom in any learning from demonstration system is that the better the demonstrations, the better the solution. For instance \cite{james2017transferring} use a programmed controller in the virtual world. Our experiments show that this is not always the case. 

As the objective of our project was to develop methods through which disabled users can demonstrate their preferred way to perform activities of daily living, in our experiments demonstrators were done by imperfect human demonstrators. The demonstrations contained errors such as failure to grasp, bumping the arm into shelves, dropping the object from the side of the shelf, overpushing, underpushing, and accidentally rotating the pushed object. In all cases, the demonstrator tried to correct the error by retrying, or through compensatory actions. In some cases, the demonstrator did not manage to complete the task. All these demonstrations, even the failed ones, were included in the training data. 

In the behavior of the trained robot, both in the virtual and the physical ones show that the robot learned to correct its mistakes. See, for instance, the experiments shown in the video\footnote{\url{https://youtu.be/9vYlIG2ozaM}} from about 2:30. This is an obvious performance improving factor, especially in the physical world, where the robot inevitably makes more errors. In fact, arguably, the robot would rarely be able to complete a task without this self-correcting behavior.

\section{Conclusions and future work}
\label{sec:Conclusions}

In this paper we have developed a technique through which a robotic arm can be taught to perform certain manipulation tasks. We focused on two tasks that are frequently required of robots that assist disabled users in activities of daily living: pick and place and push to desired pose. As disabled users can not generate large numbers of demonstrations on physical robots, we designed an approach where the user demonstrates the task in a virtual environment. These virtual demonstrations are used to teach a deep neural network based robot controller. Then, the controller is transferred to the physical robot. We found that the best performance was obtained using a network with LSTM layers and a mixture density network based error signal. We also found that having imperfect demonstrations, where users occasionally make mistakes but correct them, allows the controller to correct the inevitable mistakes it makes when transferred to a physical environment. 

Our team is working to improve these results along several directions. Multi-task learning might reduce the necessary number of demonstrations as many features are likely shared between tasks. We are extending the tasks to more complex, multi-step tasks involving multiple objects. We are working on acquiring demonstrations from multiple disabled users, who only demonstrate each task only two or three times. We are also extending the controller to an end-to-end, vision-to-control model.


\noindent{\bf Acknowledgments:} We would like to gratefully thank Sergey Levine, Navid Kardan, and Amirhossein Jabalameli for their helpful comments and suggestions. This work had been supported by the
National Science Foundation under Grant Number IIS-1409823.
 
\bibliography{refs}
\bibliographystyle{aaai}

\end{document}